# K-AID: Enhancing Pre-trained Language Models with Domain Knowledge for Question Answering

Fu Sun, Feng-Lin Li
{fusun.sf,fenglin.lfl}@alibaba-inc.com
DAMO Academy, Alibaba Group

Ruize Wang
rzwang18@fudan.edu.cn
Fudan University

Qianglong Chen
qianglong.cql@alibaba-inc.com
DAMO Academy, Alibaba Group

Xingyi Cheng
fanyin.cxy@antgroup.com
DAMO Academy, Alibaba Group

Ji Zhang
zj122146@alibaba-inc.com
DAMO Academy, Alibaba Group

## ABSTRACT

Knowledge enhanced pre-trained language models (K-PLMs) are shown to be effective for many public tasks in the literature, but few of them have been successfully applied in practice. To address this problem, we propose **K-AID**, a systematic approach that includes a low-cost **k**nowledge **a**cquisition process for acquiring domain knowledge, an effective knowledge **i**nfusion module for improving model performance, and a knowledge **d**istillation component for reducing model size and deploying K-PLMs on resource-restricted devices (e.g., CPU) for real-world application. Importantly, instead of capturing entity knowledge like the majority of existing K-PLMs, our approach captures relational knowledge, which contributes to better improving sentence level text classification and text matching tasks that play a key role in question answering (QA). We conducted a set of experiments on five text classification tasks and three text matching tasks from three domains, namely E-commerce, Government and Film&TV, and performed online A/B tests in E-commerce. Experimental results show that our approach is able to achieve substantial improvement on sentence level question answering tasks and bring beneficial business value in industrial settings.

## 1 INTRODUCTION

Pre-trained language models (PLMs) such as BERT [4], GPT [17], and RoBERTa [14] have achieved state-of-the-art performance on many NLP tasks. As PLMs capture only a general language representation learned from large-scale corpora, many efforts, such as ERNIE$_{Tsinghua}$ [31], K-BERT [13] and K-Adapter [23], have been made for injecting knowledge into PLMs for further improvement.

Knowledge enhanced PLMs are shown to be effective for many public tasks in the literature. However, such techniques can not be directly and effectively applied to domain-specific tasks in practice. Existing K-PLMs usually adopt off-the-shelf common knowledge graphs (e.g., ConceptNET [19], HowNet [16]) as knowledge sources to enhance publicly released PLMs such as BERT or RoBERTa, as in Figure 1 (a). However, this is not the case in reality. This is because that we often lack off-the-shelf domain knowledge graphs for a specific domain, where the effect of injecting general knowledge can be rather limited. Further, as shown in Figure 1 (b), the baseline PLMs are often further trained with domain corpus in the industry community, and the effect of injecting general knowledge in such situations can be largely reduced or even eliminated. Our experiments on five text classification tasks in E-commerce show that ConceptNET can help to narrowly improve the average $F_1$ of StructBERT$_{LARGE}$ [24] by 0.55%, which will be decreased to only 0.08% (almost zero) if StructBERT$_{LARGE}$ is further trained with domain corpus.

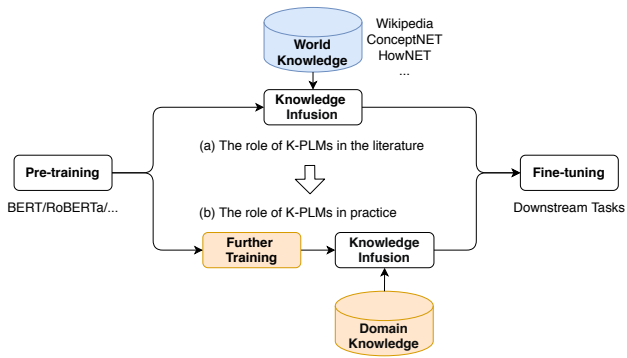

**Figure 1: The role of K-PLMs in the paradigm of pre-training and fine-tuning.**

Second, current K-PLMs mainly focus on utilizing entity related knowledge to improve token level tasks such as word in context

(WiC) [15], entity typing [28], named entity recognition (NER) and relation extraction [20, 26, 31], and are less effective on sentence level tasks such as text classification and text matching, which play a key role in question answering. In our observation, the reason can be that they consider the entities in a context (e.g., a sentence) in isolation, and usually neglect the relations thereof.

Last but not least, it is too heavy and costly to directly launch K-PLMs for real-world application due to their large scale parameters. To control cost, we need to reduce model size and accelerate inference while still maintaining accuracy.

To address the aforementioned challenges, we propose **K-AID**, a systematic approach that includes a low-cost knowledge acquisition process, an effective knowledge infusion module, and a knowledge distillation component. Unlike current K-PLMs that merely focus on models, we believe that knowledge acquisition and distillation are also of key importance to applying K-PLMs in practice. We assess the effectiveness of K-AID through a set of experiments and demonstrate its value through online A/B tests.

Our paper makes the following contributions:

- We propose K-AID, a systematic knowledge enhancement approach that consists of knowledge acquisition, infusion and distillation, for injecting domain knowledge into PLMs for further improvement and real-world application.
- We propose a relational knowledge representation that is designed for enhancing sentence level question answering tasks, and present a corresponding lightweight knowledge acquisition process that is much more cheaper than traditional KG construction.
- We demonstrate the effectiveness of K-AID through a set of experiments on classification and matching tasks in three domains, namely E-commerce, Government and Film&TV, and show its value through online A/B tests in E-commerce.

The rest of the paper is structured as follows: Section 2 introduces our approach; Section 3 describes experiments and A/B tests; Section 4 discusses example case studies; Section 5 reviews related work, and Section 6 concludes the paper and sketches directions for future work.

## 2 K-AID

To enrich pre-trained LMs with domain knowledge, we first define what domain knowledge is and specify its representation, and then divide the problem into three key steps: 1) how to acquire domain knowledge; 2) how to inject domain knowledge into pre-trained LMs; and 3) how to reduce the size of a knowledge enhanced PLM for low-cost deployment in practice.

We show an overview of K-AID in Figure 2. In general, we first acquire domain knowledge from large-scale domain corpus through mining, then integrate such knowledge into PLMs through pre-training with relation extraction, and at last distill the knowledge in a knowledge enhanced PLM to a lightweight student model to educe model size and accelerate inference.

### 2.1 Domain Knowledge

We treat knowledge as triples $(e_h, r, e_t)$ that consist of head entity, relation and tail entity, and associated sentences $d$ that describe triples. For example, the associated sentence "our school only accept Yuan Tong express delivery" captures the "specified_express_delivery" relation between "school" and "Yuan Tong express delivery", which is a piece of knowledge conveyed by the compact triple "(school, specified_express_delivery, Yuan Tong express delivery)".

Note that triples in the form of $(e_h, r, e_t)$ can also be denoted as binary relations $r(e_h, e_t)$, and then be easily extended to n-ary ones. Since we observed that many sentences contain less than or more than two entities, we accordingly extend our knowledge representation to Equation 1, which indicates that a set of entities $(e_1, e_2, ...e_n)$ reflects the relation $r$ in the context of $d$.

$$r(e_1, e_2, ...e_n) : d, n \geq 1 \quad (1)$$

This kind of knowledge is relational as it captures not only the entities in a context (e.g., sentence) but also the relations thereof. We treat such kind of knowledge in a specific domain as **domain knowledge**. We show some examples in Table 1, where the five lines illustrate three different kinds of tuples and their associated sentences. For example, "*size_selection* (child, clothes)" is a 2-ary tuple, which is expressed through the associated sentence "What size of clothes should a child wear".

### 2.2 Knowledge Acquisition

The construction of knowledge graph is fairly expensive and time-consuming because of its complexity and high accuracy requirement. To reduce the degree of difficulty, we propose to relax accuracy and simplify construction process. This makes sense because our knowledge graph is built for models, which are able to learn from noisy or weakly supervised data, instead of end users, who have very low mistake tolerance.

We show an overview of our knowledge acquisition process in Figure 3, where we take domain phrases as entities, and treat class labels as relations. We start with a domain corpus $D_C$ consists of $m$ sentences. To identify entities from sentences, we first employ automated phrase mining [11, 18] to extract from $D_C$ a set of quality phrases (e.g., "child" and "clothes") that exceed a certain frequency $f$ and quality score $\tau$ [1] (step ①). Treating the phrases as domain specific entities $E$, we next use dictionary matching to detect entity mentions[2] from each sentence as in step ② (e.g., "$d_2$: child, clothes").

To extract relations of entities, we assume that if a sentence $d$ with entities $(e_1, e_2, ..., e_n)$ belongs to a class $r$, then the entities $(e_1, e_2, ..., e_n)$ reflect the relation $r$ in the context of $d$. To assign a class label to each sentence in $D_C$, we adopt clustering to obtain a set of labels with associated sentences, and train a corresponding classifier. Specifically, we perform clustering on a randomly sampled subset of the domain corpus, denoted as $D_C^{'}$, through K-means with tf-idf vectorization [3]. After that, we ask domain experts to name the most frequent $K$ (e.g., 100) clusters based on their representative sentences (step ③). We treat clusters as classes and use labeled class names as relations $R$ (e.g., "size_selection" and "wrong_delivery"). Therefore, we are able to train a $K$-class classifier with the aforementioned $K$ clusters as training data, and use it to predict the relation of each sentence in the corpus $D_C$ (step

---
[1]In phrase mining, we use existing high-quality phrases as positive examples and the remaining phrases as negative examples to train a quality evaluation model.
[2]For simplicity, we literally equate mentions with entities.
[3]One can also use distributional representation such as average of word embedding or BERT token embedding, but tf-idf is simple and works well in our case.

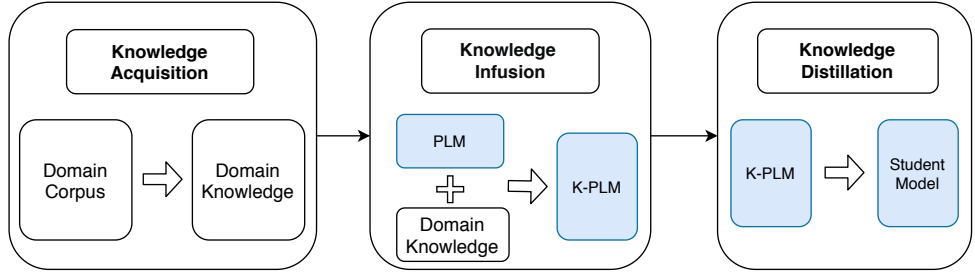

Figure 2: An overview of our K-AID approach.

Table 1: Example pieces of domain knowledge in E-commerce.

| Entities | Relation | Tuples | Tuple Type | Associated Sentence |
|---|---|---|---|---|
| shipped | reminder | reminder (shipped) | 1-ary | Haven't shipped yet? |
| child, clothes | size_selection | size_selection (child, clothes) | 2-ary | What size of clothes should a child wear |
| pants, wrong | wrong_delivery | wrong_delivery (pants, wrong) | 2-ary | You send me the wrong pants |
| school, Yuan Tong | specified_express_delivery | specified_express_delivery (school, Yuan Tong) | 2-ary | Our school only accepts Yuan Tong express delivery |
| height, weight, size | size_selection | size_selection (height, weight, size) | 3-ary | With a height of 173 and a weight of 60, what size should I choose? |

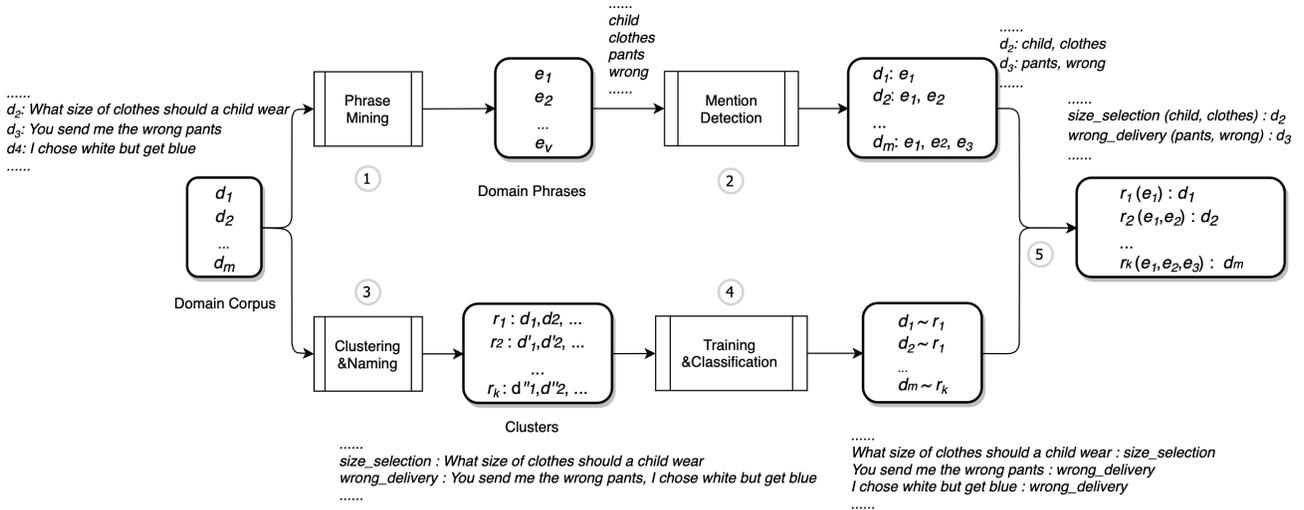

Figure 3: An overview of knowledge mining.

④). In this way, we are able to know both the entities in and the relation of each sentence, and accordingly collect a large number of tuples and associated texts (step ⑤).

Our approach is a simplification of traditional KG construction as summarized in [27]: we omitted many tasks such as schema definition, entity typing and quality checking. The only part that needs human review in our process is the naming of dozens of or hundreds of clusters based on representatives, and the check of training data for the relation classifier, while all the other parts can be automated. Even so, we can further suggest labels for clusters based on extracted domain phrases, boost training data collection through data augmentation (e.g., self-training [1]), to further improve productivity. In general, our approach takes only a few days (e.g, 1 ∼ 3) to build a light KG for a given domain, which is much more cheaper and convenient than traditional ones.

## 2.3 Knowledge Infusion

The overview of our knowledge infusion module is shown in Figure 4. Based on our knowledge representation, we adopt relation classification as the pre-training task to learn from the domain knowledge mined from a given domain corpus. Specifically, as shown in Figure 4 (a), we take a sentence as input, and use the concatenation of "[CLS]" embedding and averaged entity embedding to predict the target relation.

We follow K-Adapter [23] to inject the mined knowledge into adapters, in which way we are able to handle the tasks of different domains through training domain-specific adapters. As shown in Figure 4 (b), the backbone of our model is a pre-trained language model, enhanced by an adapter consists of multiple Transformer [22] layers. To handle arbitrary number of entities in a sentence (e.g., 0, 1, 2 or more), we use an extra array of indexes to mark the start token position of each entity. Moreover, considering that there can be no entity in a sentence, in which situation the average of entity embedding is a meaningless zero vector, we add a placeholder "[PLC]" just after "[CLS]" in each input sentence and give the placeholder an entity mark of 1. For example, the input in Figure 4 in fact becomes "[CLS] [PLC] our scholl only accepts Yuan Tong express [SEP]".

We assume that each sentence is a sequence of $L$ tokens $x = [x_1, x_2, ..., x_L]$ and comes with a list of identified entities $e = [e_{PLC}, e_1, e_2, ..., e_n]$, where $e = [e_{PLC}]$ if no entity identified and $e_{PLC}$ is inserted as a placeholder, each $e_i$ is defined by its start token index $s_{e_i}$. We formulate our model as Equation 2.

$$\begin{aligned}
X_0 &= Token\_Embedding(x) \\
X_1 &= PLM(X_0, n\_layers = L_1) \\
X_2 &= Adapter(X_1, n\_layers = L_2, layers = l_{f(j)}) \\
X_3 &= Fusion(X_1, X_2) \\
X_4 &= Pooling(X_3, position = \{s_{e_0}, s_{e_i}\}) \\
X_5 &= Projection(X_4) \\
\widetilde{Y} &= Softmax(X_5)
\end{aligned} \quad (2)$$

Token embedding $X_0$ follows the BERT convention for an input sequence of tokens $x$. $X_1$ is the holistic output[4] of the backbone PLM, which consists of $L_1 = 24$ Transformer layers, and $X_2$ is the final output of adapter that is composed of $L_2 = 3$ adapter layers. Each adapter layer $l_j$ takes as input the output of the adapter layer $l_j - 1$ (the input is $\mathbf{0}$ if $l_j - 1 \leq 0$) and that of the PLM layer $l_{f(j)}$ ($0 \leq f(j) \leq L_1 - 1$). After that, we obtain $X_3$ through concatenating the last layer embedding of $X_1$ and $X_2$, and form the sentence embedding $X_4$ through concatenating the "[CLS]" and averaged entity embedding. At last, we project $X_4$ to classification space $X_5$ and perform softmax to get the prediction $\widetilde{Y}$. The model is trained through using the cross entropy loss function.

The difference between our model and K-Adapter can be detailed through the calculation of $X_0$ and $X_4$ in Equation 2. Specifically, we remove the special tokens added before and after a certain entity in input layer and perform pooling using the embedding of the first token of each entity. In this way, our model is able to handle $n$-ary tuples, and perform consistently at both pre-training and fine-tuning stages. Meanwhile, K-Adapter is able to deal with only two entities, and has the problem of inconsistency between pre-training and fine-tuning: it uses the pooling of two entities during pre-training while adopting "[CLS]" for fine-tuning.

## 2.4 Knowledge Distillation

Knowledge enhanced PLMs are often too heavy and costly to be directly launched for real-world application, hence we employ model distillation to reduce model size and accelerate inference while maintaining accuracy. Specifically, we follow [8] to distill the knowledge in a teacher model $M_t$ to a student model $M_s$, as shown in Equation 3, where $p$ refers to the distribution generated by the teacher model (i.e., $X_5$ in Section 2.3), $q$ represents the distribution approximated by student, $g$ denotes one-hot golden distribution, $C$ represents the cross entropy loss, and $\lambda$ is a weight parameter that balances the importance of soft targets and hard (golden) targets.

$$Loss = \lambda C(p, q) + (1 - \lambda)C(g, q) \quad (3)$$

The overview of knowledge distillation is shown in Figure 5. Operationally, we first use the original training data $D_O$ of a given task obtain the teacher model $M_t$ through fine-tuning a PLM, and use it to generate pseudo labels for unlabelled data $D_A$. We then organize the augmented data $D = D_O \cup D_A$ in the form of "$(d, l, p)$", where $d$ represents a text sentence, $l$ indicates its label (golden for $D_O$ and pseudo for $D_A$), and $p$ refers to distribution generated by the teacher model $M_t$, and finally use $D$ to train a student model (e.g., CNN [29]).

## 3 EXPERIMENTS

### 3.1 Tasks

In practice, the majority of question answering bots are built on the basis of frequently asked questions (FAQ). In such systems, knowledge is commonly represented as standard questions with corresponding answers, customer questions are mapped to the most similar standard question through either classification or matching, and finally replied with the corresponding answer. In this work, we evaluate the effectiveness of enriching PLMs with domain knowledge mainly through text classification tasks, and also use text matching tasks for validation at last.

*3.1.1 Text classification tasks.* For text classification, we use five Chinese datasets from different sub-domains of E-commerce: namely Fresh Food, Supermarket, Logistics, Online Platform, and Hotline. The datasets are organized in the form of $(q, k)$, where $q$ represents after-sales customer questions and $k$ stands for corresponding standard questions. For example, the customer question "please help me to return the product" is associated with the standard one "how to return products", which is treated as the corresponding class label. All datasets are collected from online conversations within our customer service chatbot application [12], except Hotline that is obtained through automatic speech recognition of inbound calls.

We report the statistics of these datasets in Table 2 and use $F_1$ as our evaluation metric. Note that each task has an UNKNOWN label that considers customer questions do not belong to any pre-defined class. In fact, the UNKNOWN label accounts for 29.98% of training data in each task on average.

*3.1.2 Text matching tasks.* For text matching, we use three Chinese datasets from different domains: Government, E-commerce, Film&TV. The datasets are organized in the form of $(q_1, q_2, s)$, where $q_1$ and $q_2$ are customer questions, and $s$ is an indicator that indicates whether the pair of questions are semantically equivalent, 1 for yes and otherwise 0. For example, "passport application processing progress" and "how to check the progress of my passport application" is a semantically equivalent pair. The Government dataset is about social security and accumulation fund, the E-commerce

---
[4]The output of each Transformer layer, instead of that of the final.

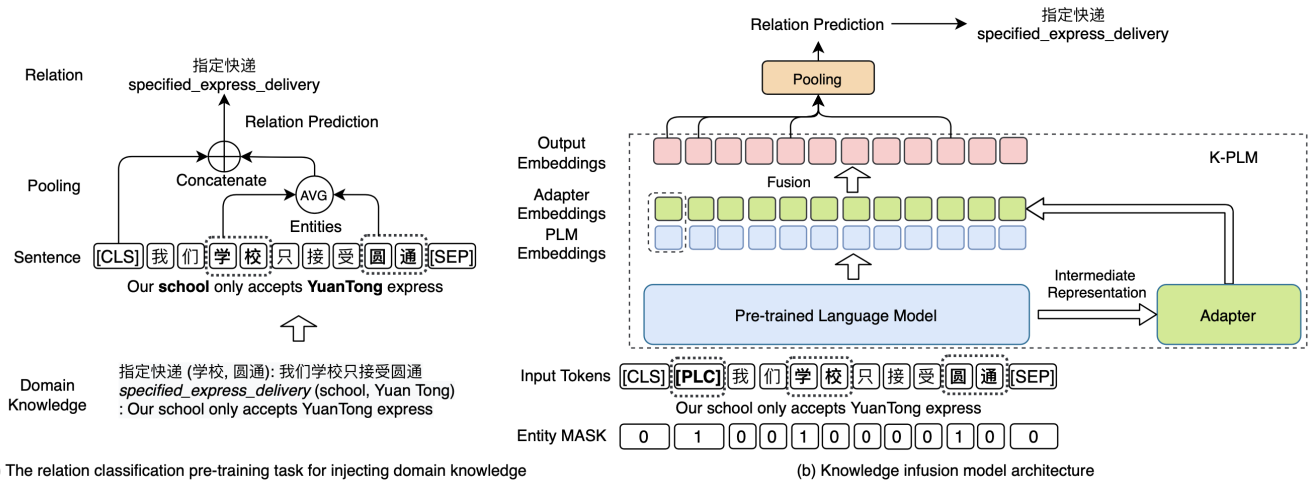

Figure 4: An overview of knowledge infusion.

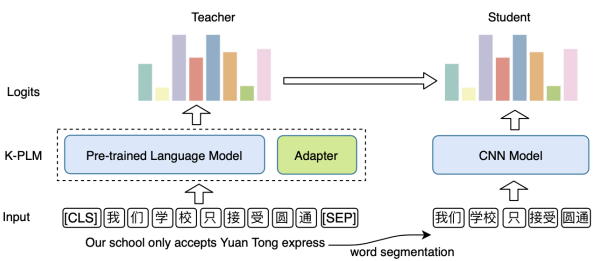

Figure 5: An overview of knowledge distillation.

Table 2: The statistics of our five text classification datasets. "#Classes" and "#CSZ" refer to number of classes, average number of training data per class, respectively.

| Dataset | #Train | #Test | #Classes | #CSZ |
|---|---|---|---|---|
| Fresh Food | 6900 | 1516 | 64 | 107 |
| Supermarket | 9320 | 2000 | 96 | 97 |
| Logistics | 5027 | 1987 | 42 | 119 |
| Online Platform | 43495 | 3500 | 91 | 477 |
| Hotline | 37589 | 4777 | 99 | 379 |

dataset is about pre-sales and after-sales, and the Film&TV dataset is about Youku membership.

We report the statistics of these datasets in Table 3 and use *AUC* as our evaluation metric. Note that the E-commerce dataset has more than 331K pairs while Film&TV has only around 5K pairs of training data, which correspond to high-resource and low-resource scenario, respectively.

Table 3: The statistics of our three text matching datasets.

| Dataset | #Train | #Test |
|---|---|---|
| Government | 31326 | 4783 |
| E-commerce | 331214 | 4478 |
| Film&TV | 4994 | 2207 |

### 3.2 Knowledge Graphs

*3.2.1 World knowledge.* We choose the Chinese version ConceptNET [19] as world knowledge as it contains both triples and associated text descriptions, other KGs such as DBPedia [2] or Freebase [3] do not fit in our case.

*3.2.2 Domain knowledge.* For domain knowledge graphs, we construct them from scratch through using the mining approach introduced in section 2.2. Moreover, we differ E-commerce KG, which captures only binary relations, from E-commerce KG++, the one that captures $n$-ary relations.

### 3.3 Baselines

*3.3.1 S-BERT.* StructBERT (S-BERT for short)[5] [24] is an extension of BERT [4] by incorporating word structural (reconstruct the correct order of shuffled words in a sentence) and sentence structural (predict both the next and previous sentence relation) into pre-training, and achieves better performance on a variety of downstream tasks than BERT. As larger models are more difficult to improve, we employ $StructBERT_{LARGE}$, instead of $StructBERT_{base}$, to evaluate the effectiveness of our approach in this work.

*3.3.2 S-BERT+Adapter+ConceptNET.* We use $StructBERT_{LARGE}$ as the backbone, inject the world knowledge captured in ConceptNET through adapter, and use this K-PLM as our baseline.

---
[5]StructBERT is a pre-trained LM produced by the DAMO Academy, Alibaba Group.

## 3.4 Experiment Settings

We denote the number of layers as $L$, the hidden size as $H$, the number of self-attention heads as $A$, the learning rate as $LR$, the batch size as $BS$, the random seed as $SD$.

*3.4.1 Further training with domain corpus.* We follow the industrial convention to further train StructBERT$_{LARGE}$ on an E-commerce conversational corpus $D_S$[6] that totals of 40GB texts with two tasks: masked language model (MLM) and session consistency (SC) task. The SC task is to justify whether a sentence in a session has been randomly replaced. We name the resulted model StructBERT$_{Domain}$ (S-BERT$_D$ for short), which is simultaneously trained on 4 Tesla P100 GPUs for around 20 days.

*3.4.2 Knowledge infusion.* We adopt StructBERT$_{LARGE}$ ($L$=24, $H$=1024, $A$=24) as the backbone, and use an adapter layer consistent with K-Adapter [23] ($L_A$=1, $H_A$=768, $A_A$=12). Specifically, we plug 3 independent adapter layers into StructBERT at layers {0, 11, 23}. Each knowledge-specific adapter (for different KG) is trained ($LR$=5e-5 and $BS$=64) on 1 NVIDIA Tesla P100 GPU with 16GB memory within 24 hours. Note that StructBERT is fixed during knowledge injection.

*3.4.3 Fine-tuning.* For fine-tuning, we run each downstream tasks 6 times ($SD$={1,42}, $LR$={2e-5, 3e-5, 5e-5}, $BS$=48) and report the average $F_1$ or $AUC$ as final result. Each task is run on 1 Tesla P100 GPU. Note that adapters are frozen and only PLMs are updated during fine-tuning.

*3.4.4 Knowledge distillation.* For knowledge distillation, we set the weight parameter $\lambda$=0.9 and adopt CNN [29] ($L$=1, $H_{word}$=300, $LR$=1e-4) as our student model. Being different from PLMs, the input of student model need to be segmented.

## 3.5 Experimental Results: Knowledge Acquisition

We constructed a KG from an E-commerce corpus $D_C$ that consists of 2.88M sentences in around three man-days by following the process introduced in Section 2.2. We set the frequency $f \geq 3$ and quality score $\tau \geq 0.5$.

We show the statistics of KGs in Table 4, where the number of sentences is larger than that of tuples as a tuple can be associated with more than one sentence. Note that E-commerce KG captures only binary relations while E-commerce KG++ captures $n$-ary ones. Moreover, our spot check shows that the acceptance rate of automatically extracted domain phrases reaches 80% and the precision of relation classification is 79.77%.

Table 4: The statistics of used knowledge graphs.

| KG | #Entities | #Relations | #Tuples | #Sentences |
|---|---|---|---|---|
| ConceptNET$_{CN}$ | 124,611 | 15 | 354,618 | 362,068 |
| E-commerce KG | 12,470 | 160 | 168,850 | 453,110 |
| E-commerce KG ++ | 12,470 | 160 | 859,845 | 1,200,000 |

---

[6]Note that the corpus ($D_S$, a set of sessions) used here is different from that ($D_C$, a set of sentences) we used to acquire domain knowledge.

## 3.6 Experimental Results: Knowledge Infusion

*3.6.1 The effect of injecting world knowledge vs. domain knowledge.* We first evaluated the effect of ConceptNET and E-commerce KG on the aforementioned text classification tasks. In this experiment, we use StructBERT$_{LARGE}$ as the backbone. As shown in Table 5, E-commerce KG outperforms ConceptNET on all the five tasks, which clearly indicates the advantage of injecting domain knowledge. On average, E-commerce KG is able to improve the average $F_1$ performance by 1.34% (80.37% → 81.71%), compared to 0.55% (80.37% → 80.92%) brought about by ConceptNET.

*3.6.2 The effect of using relation extraction with adapter.* The second experiment is to verify the value of KG construction through evaluating the effect of relation extraction and classification: if classification can be better, then there is no need for entity identification and subsequent relation extraction. To this end, we form a text classification dataset using the 453K text sentences along with their 160 attached relations in the E-commerce KG, and directly use it for knowledge infusion.

The results are shown in Table 6. We can see from the upper part that classification do performs better than relation extraction as a pre-training task for StructBERT. As one can also observe that relation extraction is in fact useful (80.37% → 80.90%), we take that it may have model parameters shifted while injecting knowledge into StructBERT due to the inconsistency between pre-training with relation extraction and fine-tuning with classification. In fact, if we inject knowledge into adapter instead of StructBERT (i.e., keep StructBERT fixed), relation extraction can help to achieve the best performance, as shown in the lower part of Table 6. In general, this experiment indicates that injecting knowledge into adapter through relation extraction is a better way, which demonstrates the value of our lightweight KG construction.

*3.6.3 The effect of injecting domain knowledge after further training.* As further training of PLMs with domain corpus is widely used in reality, it would be interesting to see the effectiveness of domain KG in such a more difficult setting. We evaluate ConceptNET and E-commerce KG on the further trained StructBERT$_{Domain}$, and report the results in Table 7. We can clearly see that further training with domain corpus can help to largely improve the average $F_1$ by 2.31% (80.37% → 82.68%). In this situation, the effect of injecting ConceptNET knowledge becomes marginal 0.08% (82.68% → 82.76%), while E-commerce KG is still able to improve the average $F_1$ by 0.36% (82.68% → 83.04%). This indicates domain knowledge can still be effective even if PLMs are further trained with domain corpus.

*3.6.4 The effect of capturing n-ary relational knowledge.* To explore the effect of $n$-ary relational knowledge, we randomly sample roughly the same number of training examples that have more than one entity from E-commerce KG++, and use it to pre-train another adapter. As we can see from the results reported in Table 8, $n$-ary relational knowledge is able to further improve the performance of four out of the five classification tasks, and increase the average $F_1$ by 0.14% (83.04% → 83.18%).

*3.6.5 Validation on text matching tasks.* We also evaluated the effectiveness of injecting general knowledge versus domain knowledge on the three text matching tasks[7] and show the results in Table 9. Similarly, we can observe that injecting ConceptNET knowledge is hard to improve the performance of StructBERT$_{Domain}$ (slightly decreased: 83.32% → 83.23%), which is further pre-trained on domain corpus, while injecting domain knowledge is still able to further improve the average $AUC$ by 0.74% (83.32% → 84.06%). Moreover, as expected, our approach is able to improve more on low-resource task (e.g., 1.02% = 85.73% - 84.72% for Film&TV) and less on high-resource task (e.g., 0.24% = 87.46% - 87.22% for E-commerce) when compared with StructBERT$_{Domain}$, the harder baseline.

Table 5: The effect of injecting general knowledge versus domain knowledge ($F_1$).

| Model | Fresh Food | Supermarket | Logistics | Online Platform | Hotline | Avg. |
|---|---|---|---|---|---|---|
| S-BERT | 89.12% | 67.49% | 80.31% | 86.71% | 78.25% | 80.37% |
| S-BERT+Adapter+ConceptNET | 88.74% | 68.41% | 80.96% | 87.09% | 79.41% | 80.92% |
| S-BERT+Adapter+E-commerce KG | **89.25%** | **70.36%** | **81.30%** | **87.10%** | **79.46%** | **81.71%** |

Table 6: The effect of using relation extraction with adapter on E-commerce KG ($F_1$). C: classification; RE: relation extraction.

| Model | Fresh Food | Supermarket | Logistics | Online Platform | Hotline | Avg. |
|---|---|---|---|---|---|---|
| S-BERT | 89.12% | 67.49% | 80.31% | 86.71% | 78.25% | 80.37% |
| S-BERT (C) | 89.02% | **70.52%** | 80.52% | 86.98% | 79.17% | 81.24% |
| S-BERT (RE) | 88.62% | 70.08% | 79.97% | 86.82% | 79.02% | 80.90% |
| S-BERT+Adapter (C) | 89.24% | 69.94% | 81.20% | 86.97% | 79.32% | 81.33% |
| S-BERT+Adapter (RE) | **89.25%** | 70.36% | **81.30%** | **87.10%** | **79.46%** | **81.71%** |

Table 7: The effect of injecting domain knowledge after further training PLMs with domain corpus ($F_1$).

| Model | Fresh Food | Supermarket | Logistics | Online Platform | Hotline | Avg. |
|---|---|---|---|---|---|---|
| S-BERT | 89.12% | 67.49% | 80.31% | 86.71% | 78.25% | 80.37% |
| S-BERT$_D$ | 90.37% | 71.40% | 83.98% | 87.34% | **80.31%** | 82.68% |
| S-BERT$_D$+Adapter+ConceptNET | **90.57%** | 71.12% | 84.20% | 87.61% | 80.29% | 82.76% |
| S-BERT$_D$+Adapter+E-commerce KG | 90.56% | **72.18%** | **84.45%** | **87.76%** | 80.24% | **83.04%** |

Table 8: The effect of capturing $n$-ary relational knowledge ($F_1$).

| Model | Fresh Food | Supermarket | Logistics | Online Platform | Hotline | Avg. |
|---|---|---|---|---|---|---|
| S-BERT$_D$+Adapter+E-commerce KG | 90.56% | 72.18% | **84.45%** | 87.76% | 80.24% | 83.04% |
| S-BERT$_D$+Adapter+E-commerce KG++ | **90.68%** | **72.51%** | 84.37% | **87.83%** | **80.49%** | **83.18%** |

### 3.7 Experimental Results: Knowledge Distillation

To launch knowledge enhanced PLMs for real-world application, we distill the knowledge in a K-PLM into a lightweight student CNN model by following the procedure described in Section 2.4. To ensure performance, we use the best teacher model among the 6 candidates of each task for distillation. Moreover, we augment each task with another 200K training data through collecting unlabelled customer questions from each sub-domain, and generating pseudo labels for them using the corresponding teacher model.

We show the distillation results in Table 11, from which we can see that the average performance of student models on the five text classification tasks is only 1% lower than that of teacher models for both StructBERT$_{Domain}$ and StructBERT$_{Domain}$ with adapter.

We also analyzed the model size of both teacher and student models, and report the statistics in Table 10. One can see that the adapter in our approach only introduces an extra set of 45M parameters, meaning that the pre-training of an adapter is much faster than that of a backbone PLM. Moreover, the size of student model is much smaller than that of teachers, indicating that the student model can be deployed on restricted resources (e.g., CPU) at a low cost. Our experiments show that the inference speed of student model is much faster: a 139X speedup for StructBERT$_{Domain}$ and 190X speedup for StructBERT$_{Domain}$ with adapter on average.

Table 10: The size of teacher and student models.

| Model | #Parameters |
|---|---|
| S-BERT$_D$ | 340M |
| S-BERT$_D$+Adapter+E-commerce KG | 340M+45M |
| Student CNN (excluding word embedding) | 7M |

### 3.8 Online A/B Tests

We currently have launched three out of the five students models online for real-world application, namely Fresh Food, Logistics and Online Platform, and conducted A/B tests on the distilled models through equally distributing customer questions. We found that if we keep the precision consistent, student models distilled from StructBERT$_{Domain}$ with adapter are able to improve the recall by 3.189% (i.e., addressing more questions with the same precision) and increase the degree of customer satisfaction by 1.43%, on average. The other two are in progress.

---
[7]To perform the matching experiments on Government and Film&TV, we have built another two lightweight KGs and accordingly trained two adapters.

Table 9: The effect of injecting general knowledge versus domain knowledge on text matching tasks (*AUC*).

| Model | Government | E-commerce | Film&TV | avg. |
|---|---|---|---|---|
| S-BERT | 77.66% | 87.15% | 83.53% | 82.78% |
| S-BERT$_D$ | 78.02% | 87.22% | 84.72% | 83.32% |
| S-BERT$_D$+Adapter+ConceptNET | 78.25% | 86.60% | 84.84% | 83.23% |
| S-BERT$_D$+Adapter+E-commerce KG | **79.00**% | **87.46**% | **85.73**% | **84.06**% |

Table 11: The result of knowledge distillation ($F_1$).

| Model | Fresh Food | Supermarket | Logistics | Online Platform | Hotline | Avg. |
|---|---|---|---|---|---|---|
| S-BERT$_D$ | 90.95% | 71.83% | 84.20% | 87.66% | 80.55% | 83.04% |
| S-BERT$_D$+Adapter+E-commerce KG | 91.30% | 72.90% | 84.66% | 88.08% | 80.94% | 83.58% |
| S-BERT$_D$ + Distillation | 90.20% | 71.04% | 83.05% | 86.28% | 79.52% | 82.02% |
| S-BERT$_D$+Adapter+E-commerce KG+Distillation | 90.83% | 72.23% | 83.31% | 86.53% | 80.01% | 82.58% |

## 4 CASE STUDY

In this section, we use real examples in Fresh Food to demonstrate how our model is able to enhance PLM with knowledge. As shown in Table 12, for the example "Please place my package into the Hive Box at my front door", the golden label of which is "deliver method", the top-3 predictions of StructBERT$_{Domain}$ and our K-PLM are ['UNKNOWN', 'deliver method', 'specified delivery'] and ['deliver method', 'UNKNOWN', 'specified delivery'], respectively. We can see that StructBERT$_{Domain}$ is prone to confuse the correct label with UNKNOWN, which account for a considerable portion of training data in each task and may semantically interweave with other classes. As expected, with integrated knowledge, our model is able to alleviate such confusion and predict the correct answer.

We further analyzed the predicted logits of each testing example, and found that the average of normalized maximal logits increases from 0.891 to 0.946, and the average of normalized variance increases from 0.012 to 0.014. This reflects that the model becomes more confident about its predictions after knowledge infusion.

## 5 RELATED WORK

In recent years, there is an emerging branch of incorporating structured knowledge into PLMs for further improvement. A popular approach is to inject factual knowledge through infusing pre-trained entity embeddings (e.g., ERNIE$_{Tsinghua}$ [31], KEPLER [25], Know-BERT [15]) or incorporating symbolic knowledge triples (e.g., K-BERT [13]). The main problem of using pre-trained entity embeddings, as pointed out by CoLAKE [20], is the separation between knowledge embedding and language embedding, and the missing of rich (explicit) contextual information of an entity in a KG. Although the incorporating of knowledge triples in K-BERT [13] alleviates the aforementioned separation problem, we observed that it is rather limited on improving domain-specific question answering tasks with general knowledge. Being different from these methods, we focus on addressing practical domain-specific tasks through acquiring and infusing domain knowledge. Moreover, the capturing of knowledge as tuples with associated sentences, and the injecting of knowledge through relation extraction and adapter, contribute to alleviate the isolated representation of knowledge and language.

We performed an investigation on the knowledge sources and downstream tasks of K-PLMs, and show an overview in Table 13[8]. We can see from the table that the majority of methods are designed for knowledge-driven tasks, more specifically, entity-centric tasks, including sequence labeling (e.g., NER, POS), entity typing, relation classification, and entity-centric question answering (most of the answers are entities). Only a few approaches are evaluated on text classification and/or text matching. As current approaches inject knowledge into PLMs mainly through enhancing entity representation and fail to capture the relation between them within a context, we observed that sentence level tasks, which are more important in question answering, gain less improvement than token level ones. This observation is similar to that from ZEN [5]. As such, in this work we lay emphasis on sentence level text classification and matching tasks, and propose to improve their performances through incorporating relational knowledge.

In addition, many of the existing methods need to update the original parameters of PLMs when injecting knowledge, which requires retraining the entire K-PLM for each type of knowledge and probably leads to catastrophic forgetting. K-Adapter [23] proposes to address this problem by incorporating knowledge into adapters that are plug-ins of PLMs. We follow K-Adapter to inject knowledge into adapters, enjoying the benefit of easy training and domain adaptation through training different adapters.

Unlike previous work, we aim to explore the practical application of K-PLMs in specific domains, where world knowledge are less effective. We consider domain knowledge acquisition, infusion and distillation for a systematic approach, and fully exploit industrial datasets to examine the effect of K-PLMs in real scenarios.

## 6 CONCLUSION

In this paper, we present K-AID, a systematic approach that includes knowledge acquisition, infusion and distillation, to enhance pore-trained LMs with domain knowledge for question answering. Our approach alleviates the key challenges when applying K-PLMs in practices: the lack of domain knowledge, the effectiveness of knowledge infusion, and the feasibility of launching for real-world application. We demonstrate its effectiveness through a set of experiments and online A/B tests: (a) domain knowledge is more effective than general knowledge, and still useful even if the adopted PLM is adequately further trained with domain corpus[9]; (b) achieving substantial improvement on sentence-level classification and matching tasks, which play a key role in question answering; (c) serving online through distillation and bringing beneficial business value in industrial settings. More importantly, our approach is domain-independent and can be easily applied to any other domain that has a corpus.

---
[8]We do not consider lexical knowledge (e.g., LIBERT [9], Sense-BERT [10]) or syntax knowledge in this paper.

Table 12: Case study examples in the Fresh Food sub-domain of E-commerce.

| Examples | True Label | Model | Predicted Labels (Top-3) | Predicted Logits (Top-3) |
| --- | --- | --- | --- | --- |
| Please place my package into the Hive Box at my front door. | deliver method | S-BERT$_D$ | ['UNKNOWN', 'deliver method', 'specified delivery'] | [10.014635, 8.579499, 5.4292364] |
| | | Ours | ['deliver method', 'UNKNOWN', 'specified delivery'] | [10.259869, 9.321176, 7.377519] |
| I need to trouble shopkeeper to select a better one. | checking remind on shipments | S-BERT$_D$ | ['UNKNOWN', 'check remind on shipments', 'change items'] | [9.189705, 6.5492687, 5.072327] |
| | | Ours | ['check remind on shipments', 'UNKNOWN', 'gift packaging'] | [12.580814, 9.592464, 5.261752] |

Table 13: An overview of investigated knowledge enhanced pre-trained language models (K-PLMs).

| Method | Knowledge Source | Downstream Task |
| --- | --- | --- |
| ERNIE$_{Tsinghua}$ [31] | Wikipedia, WikiData | Relation Classification, Entity Typing |
| WKLM [28] | Wikipedia, WikiData | Entity Typing, Entity-Centric QA |
| KEPLER [25] | Wikipedia, WikiData | Relation Classification, Entity Typing, LINK Prediction |
| KnowBERT [15] | Wikipedia, WordNet | Relation Classification, Entity Typing, Word in Context |
| CoLAKE [20] | Wikipedia | Relation Classification, Entity Typing |
| BERT-MK [7] | Unified Medical Language System (UMLS) | Relation Classification, Entity Typing |
| K-BERT [13] | HowNet, CN-DBPedia, MedicalKG | Sequence Labeling, Text Classification, Text Matching |
| EAE [6] | Wikipedia | Entity-Centric QA |
| ERNIE$_{Baidu}$ [21] | Wikepedia, Baidu Baike/News/Tieba | NLI, Text Matching, NER, Sentimental Analysis, Retrieval QA |
| E-BERT [30] | Domain Phrases, Product Graph | Sequence Labeling, Text Classification |
| ZEN [5] | N-grams, Wikipedia | Sequence Labeling, Text Classification, Text Matching, NLI |
| K-Adapter [23] | Wikipedia, WikiData, Dependency Parser Graph | Relation Classification, Entity Typing, Question Answering |

There are many interesting and challenging problems to be further explored. One is to adapt our approach for token-level tasks such as NER and entity typing. Another one is the low-resource problem in multilingual scenario, where we often lack training data or even unlabelled data for minority languages. It is very valuable if K-PLMs can help to improve the performance of low-resource multilingual tasks, because we are currently promoting our chatbot to more than 20 kinds of languages all around the world. Last but not least, it is interesting to explore more lightweight knowledge representation such as dictionary knowledge, to make knowledge infusion more efficient and easier.

---
[9]Many K-PLMs are noneffective for high-resource tasks or the backbone PLM is further trained with domain corpus.